\documentclass[10pt,twocolumn,letterpaper]{article}

\usepackage[pagenumbers]{cvpr} % To force page numbers, e.g. for an arXiv version

\usepackage{graphicx}
\usepackage{amsmath}
\usepackage{amssymb}
\usepackage{booktabs}
\usepackage{algorithm}
\usepackage{algorithmic}
\usepackage{latexsym}
\usepackage{multirow}
\usepackage{multicol}

\usepackage[pagebackref,breaklinks,colorlinks]{hyperref}

\usepackage[capitalize]{cleveref}
\crefname{section}{Sec.}{Secs.}
\Crefname{section}{Section}{Sections}
\Crefname{table}{Table}{Tables}
\crefname{table}{Tab.}{Tabs.}

\def\cvprPaperID{8711} % *** Enter the CVPR Paper ID here
\def\confName{CVPR}
\def\confYear{2022}

\begin{document}

\title{ART-Point: Improving Rotation Robustness of Point Cloud Classifiers via Adversarial Rotation}

\author{Robin Wang\textsuperscript{\rm 1},
        Yibo Yang\textsuperscript{\rm 2,}\thanks{Corresponding author.}\ ,
        Dacheng Tao\textsuperscript{\rm 2}\\
        \textsuperscript{\rm 1} Key Laboratory of Machine Perception (MOE), School of Artificial Intelligence, Peking University\\
        \textsuperscript{\rm 2} JD Explore Academy, China\\
{\tt\small \{robin\_wang, ibo\}@pku.edu.cn}, {\tt\small dacheng.tao@gmail.com}
}

\maketitle

%%%%%%%%% ABSTRACT
\begin{abstract}
Point cloud classifiers with rotation robustness have been widely discussed in the 3D deep learning community. Most proposed methods either use rotation invariant descriptors as inputs or try to design rotation equivariant networks. However, robust models generated by these methods have limited performance under clean aligned datasets due to modifications on the original classifiers or input space. In this study, for the first time, we show that the rotation robustness of point cloud classifiers can also be acquired via adversarial training with better performance on both rotated and clean datasets. Specifically, our proposed framework named ART-Point regards the rotation of the point cloud as an attack and improves rotation robustness by training the classifier on inputs with Adversarial RoTations. We contribute an axis-wise rotation attack that uses back-propagated gradients of the pre-trained model to effectively find the adversarial rotations. To avoid model over-fitting on adversarial inputs, we construct rotation pools that leverage the transferability of adversarial rotations among samples to increase the diversity of training data. Moreover, we propose a fast one-step optimization to efficiently reach the final robust model. Experiments show that our proposed rotation attack achieves a high success rate and ART-Point can be used on most existing classifiers to improve the rotation robustness while showing better performance on clean datasets than state-of-the-art methods.

\end{abstract}

%%%%%%%%% BODY TEXT
\section{Introduction}

A very basic requirement for point cloud classification is expecting the network to obtain stable predictions on inputs undergoing rigid transformations since such transformations do not change the shape of the object, let alone change its semantic meanings. This basic requirement is even more important in practical applications. For example, when a robot is identifying and picking up an object, the object is usually in an unknown pose. However, many studies \cite{zhao2020isometry,deng2021vector, li2021rotation} have shown that most existing point cloud classifiers can be easily attacked by simply rotating the inputs. To use these classifiers we require to align all input objects which is a very expensive and time-consuming process. To this end, how to improve the robustness of point cloud classifiers to arbitrary rotations, becomes a very popular and necessary research topic.

In order to make the network robust to rotated inputs, most existing works can be classified into three categories:
(1) \textbf{Rotation Augmentation Methods} attempt to augment the training data using rotations and have been widely used in the earlier point cloud classifiers \cite{qi2017pointnet, qi2017pointnet++, wang2019dynamic}. However, data augmentation can hardly be applied to improve model robustness to arbitrary rotations due to the astronomical number of rotated data \cite{zhao2019rotation}. (2) \textbf{Rotation-Invariance Methods} propose to convert the input point clouds into geometric descriptors that are invariant to rotations. Typical invariant descriptors can be the distance and angles between local point pairs \cite{deng2018ppf,chen2019clusternet,zhang2020global,zhang2019rotation} or point norms \cite{li2021rotation, zhao2019rotation} and principal directions \cite{zhang2020global} calculated from global coordinates.
(3) \textbf{Rotation-Equivariance Methods} try to solve the rotation problem from the perspective of model architectures. 
For example, \cite{weiler20183d,chen2021equivariant,poulenard2021functional,thomas2018tensor} use convolution with steerable kernel bases to construct rotation-equivariant networks and \cite{deng2021vector, zhao2020quaternion,shen20203d} modify existing networks with equivariant operations.
% By expecting a rotation of the object in Euclidean space induces a rotation of the features in feature space. 
% Rotation-equivalent features are much more expressive thanks to their ability to retain information about the input group transform on the feature maps.
While both methods (2) and (3) can effectively improve model robustness to arbitrary rotations, they either require time-consuming pre-processing on inputs or need complex architectural modifications, which will result in limited performance on clean aligned datasets.

In this paper, we try to explore a new technical route for the rotation robustness problem in point clouds. Our method is inspired by adversarial training \cite{madry2017towards}, a typical defense method to improve model robustness to attacks. The idea of adversarial training is
straightforward: it augments training data with adversarial examples in each training loop. Thus adversarially trained models behave more normally when facing adversarial
examples than standardly trained models. 
% The main idea of adversarial training is to increase the robustness of a network against a certain attack by training the network on the most aggressive samples generated under this attack. 
Adversarial training has shown its great effectiveness in improving model robustness to image or text perturbations \cite{shafahi2019adversarial, xie2019intriguing,ganin2016domain,ebrahimi2017hotflip,liu2017adversarial}, while keeping a strong discriminative ability. In 3D point clouds, \cite{sun2020adversarial,liu2019extending} also successfully leverage adversarial training to defend against point cloud perturbations such as random point shifting or removing. However, using adversarial training to improve the rotation robustness of point cloud classifiers has rarely been studied. 

To this end, by regarding rotation as an attack, we develop the ART-Point framework to improve the rotation robustness by training networks on inputs with \textbf{A}dversarial \textbf{R}o\textbf{T}ations. 
Like the general framework of adversarial training, ART-Point forms a classic min-max problem, where the max step finds the most aggressive rotations, on which the min step is performed to optimize the network parameters for rotation robustness. 
For the max step, we propose an axis-wise rotation attack algorithm to find the most offensive rotating samples. Compared with the existing rotation attack algorithm \cite{zhao2020isometry} that directly optimizes the transformation matrix, our method optimizes on the rotation angles which reduces the optimization parameters, while ensuring that the attack is pure rotation to serve for the adversarial training. 
For the min step, we follow the training scheme of the original classifier to retrain the network on the adversarial samples. To overcome the problem of over-fitting on adversarial samples caused by label leaking \cite{kurakin2016adversarial}, we construct a rotation pool that leverages the transferability of adversarial rotations among point cloud samples to increase the diversity of training data.
Finally, inspired by ensemble adversarial training \cite{tramer2017ensemble}, we contribute a fast one-step optimization method to solve the min-max problems. Instead of alternately optimizing the min-max problem until the model converges, the one-step method can quickly reach the final robust model with competitive performance. 

% By training the classifier on inputs with the adversarial rotations, ART-Point attains models with retrained parameters that are robust against both randomly and adversarially rotated samples. In contrast to previous robust methods, ART-Point enjoys many advantages. 
Compared with the rotation-invariant and equivariant methods, the ART-Point framework aims to optimize network parameters such that the converged model is naturally robust to both arbitrary and adversarial rotations, without the necessity of either geometric descriptor extractions or architectural modifications that may impede the model to learn discriminative features. So our resulting robust model better inherits the original performance on the clean (aligned) datasets. It has no constraint on the model design and can be integrated on most point cloud classifiers. 

% Compared with method (1), ART-Point can find
% more effective augmentation groups
% to improve model robustness to arbitrary rotations. Compared to (2)(3), ART-Point improves rotation robustness without complex descriptor extraction or architectural modification. More importantly, the resulting robust model also better inherits the performance of the original model on clean datasets.

% 有随机旋转增强的优点，同时可以

% , we propose to use Adversarial RoTation to improve model's rotation robustness
% The idea behind adversarial training is similar to data augmentation, but there are two main differences: 
% (1) Data augmentation will be trained on multiple transformations at the same time, while adversarial training is only focus on specific attack.
% (2) Data augmentation applies random rotations on the input but the adversarial training requires to find the most aggressive rotations on certain inputs.
% These two differences make adversarial training can efficiently find most suitable transformation group than data augmentation to improve the model's robustness to infinite rotations in SO(3) group.

In experiments, we mainly verify the effectiveness of our methods under two datasets ModelNet40 \cite{wu20153d} and ShapeNet16 \cite{yi2016scalable}. We adopt PointNet \cite{qi2017pointnet}, PointNet++ \cite{qi2017pointnet++} and DGCNN \cite{wang2019dynamic} as the basic classifiers. Firstly, compared with the existing rotation attack method \cite{zhao2020isometry}, our proposed attack achieves a higher attack success rate. Then, compared with existing rotation robust classifiers, our best model (ART-DGCNN) shows a more robust performance on randomly rotated datasets. Meanwhile, our methods generally show less accuracy reduction on clean aligned datasets. Beyond arbitrary rotations, the resulting models also show a solid defense against adversarial rotations.\footnote[1]{Code address: \url{https://github.com/robinwang1/ART-Point}.}
Our contributions can be summarized as follows:
\begin{itemize}
    \item For the first time, we successfully improve the rotation robustness of point cloud classifiers from the perspective of model attack and defense. 
    Our proposed framework, ART-Point, enjoys fewer architectural modifications than previous rotation-equivariant methods and requires no descriptor extractions on input data. 
    \item We propose an axis-wise rotation attack algorithm to efficiently find the most aggressive rotated samples for adversarial training. A rotation pool is designed to avoid over-fitting of models on adversarial samples. We also contribute a fast one-step optimization to solve the min-max problem.
    \item We validate our method on two datasets with three point cloud classifiers. The results show that our attack algorithm achieves a higher attack success rate than existing methods. Moreover, the proposed ART-Point framework can effectively improve model rotation robustness allowing the model to defend against both arbitrary and adversarial rotations, while hardly affecting model performance on clean data.
\end{itemize}

% In terms of attack, we propose a direction-by-direction angle attack algorithm based on C\&W, which ensures the effectiveness of the attack and improves the offensiveness of the sample. In terms of defense, we have constructed static and dynamic attack sample pools and conducted confrontation training respectively.
% 3. We validated our method on three commonly used point cloud classifiers. Experiments show that compared with existing methods, our method can effectively improve the robustness of these models to rotation, and it can also show higher performance for ideal non-rotation samples. Theoretically, our attack-defense framework can be used to improve the robustness of the model against attacks other than rotation.

%-------------------------------------------------------------------------
\section{Related Work}
\subsection{Rotation Robust Point Cloud Classifiers}
    % Rotation robust is a desirable property for point clouds classification. 
% Existing work for improving rotation robustness of point cloud classifiers can mainly be divided into three categories: rotation augmentation method, rotation-invariant method and rotation-equivariant method.
\textbf{Rotation Augmentation.} The initial work of the point cloud classifier \cite{qi2017pointnet,qi2017pointnet++,wang2019dynamic} adopt rotation augmentation during training to improve rotation robustness. Nevertheless, rotation augmentation can only result in models robust to a small range of angles. More recently, to obtain models robust to arbitrary rotation angles, both rotation-invariance and rotation-equivariance methods are proposed.

\textbf{Rotation-invariance methods} extract rotation-invariant descriptors from point clouds as model inputs. For example, \cite{deng2018ppf,poulenard2019effective,chen2019clusternet,zhang2019rotation} cleverly construct distances and angles from local point pairs. \cite{zhang2020global,zhao2019rotation,li2021rotation} further extend local invariant descriptors with global invariant contexts. In addition to using invariant descriptors with a clear geometric meaning, \cite{rao2019spherical,poulenard2019effective,liu2018deep} also design invariant convolutions to automatically learn various descriptors for processing.

\textbf{Rotation-equivariance methods} expect the learned features to rotate correspondingly with the input thus resulting in rotation robust models. Most of these works usually rely on rotation-equivariant convolutions \cite{cohen2018spherical,weiler20183d,thomas2018tensor,esteves2018learning,poulenard2021functional,chen2021equivariant,he2021efficient} to construct equivariant networks. 
Other works like \cite{deng2021vector,zhao2020quaternion,shen20203d} attempt to modify modules in existing point cloud classifiers \cite{qi2017pointnet,qi2017pointnet++,wang2019dynamic} to make them rotation-equivariant.

However, these methods usually require specific descriptors or network modules which will reduce the performance of the classifier on the aligned datasets. Our study differs from these methods in that we try to obtain a robust model by optimizing the parameters without changing the input space or network architectures.

\subsection{Adversarial Training}
Adversarial Training \cite{goodfellow2014explaining, madry2017towards} has been proved to be the most effective technique against adversarial attacks \cite{pang2020bag,maini2020adversarial,schott2018towards}, receiving considerable attention from the research community. Unlike other defense strategies, adversarial training aims to enhance the robustness of models intrinsically \cite{bai2021recent}. 
This property makes adversarial training widely used in various fields to improve the robustness of the model, including image recognition \cite{gong2021maxup, shafahi2019adversarial, xie2019intriguing,ganin2016domain}, text classification \cite{miyato2016adversarial,ebrahimi2017hotflip,liu2017adversarial, morris2020textattack}, relation extraction \cite{wu2017adversarial} etc. 
In 3D point clouds classification, adversarial training can also be effectively used. For example, \cite{liu2019extending} employs adversarial training to improve the model robustness to point shifting perturbation by training on both clean and adversarially perturbed point clouds. 
\cite{sun2020adversarial} presents an in-depth study showing how adversarial training behaves in point cloud classification. However, existing works only focus on improving the model’s robustness to perturbations of random point shifting or removing \cite{xiang2019generating,yang2019adversarial,lang2020geometric,zheng2019pointcloud,liu2021pointguard, gong2021maxup}.
% These defenses are effective against simple attacks, they can hardly improve the robustness against rotation.

Recently, \cite{zhao2020isometry} designs a rotation attack algorithm for existing point cloud classifiers. Yet it does not provide detailed strategies to defense the rotation attack. 
As a comparison, we design a new attack algorithm that enjoys a higher attack success rate. More importantly, it serves for our adversarial training framework that generates model naturally defending against both arbitrary and adversarial rotations. 

\begin{figure*}
    \centering
    \includegraphics[scale=0.8]{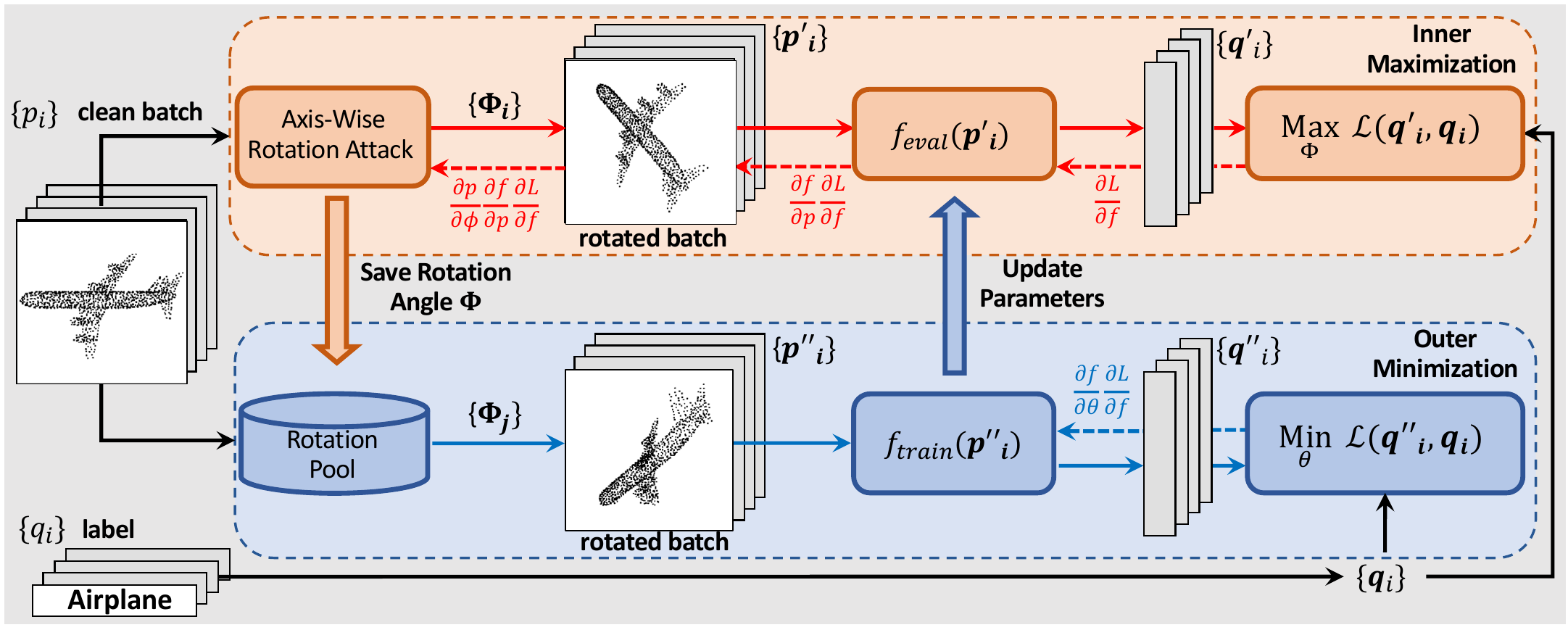}
    \caption{The general pipeline of our adversarial training approach. In the upper branch, the network takes a clean batch (aligned object) as inputs and finds the most aggressive attack angles by maximizing the classification loss of the eval model. The attack angles will be stored by class in the rotation pool. In the lower branch, the network samples angles from the rotation pool to produce adversarial point clouds for re-training the classifier to obtain the rotation robust model. The red and blue dashed lines respectively indicate routes of the backward gradient in two optimization tasks and point to the final optimized parameters. In the real implementations, the one-step optimization will construct the rotation pool by attacking multiple eval models, while the iterative optimization will update the parameter of the eval model by parameters of the latest re-trained model in each min-max iterations.}
    \label{fig1}
\end{figure*}

\section{Methods}
In this section, we first provide a brief review of adversarial training (Sect. \ref{Sect3.1}). Then, we reformulate the adversarial training objective under rotation attack of point clouds  (Sect. \ref{Sect3.2}). Next, we propose attack (Sect. \ref{Sect3.3}) and defense (Sect. \ref{Sect3.4}) algorithms to obtain good solutions to the reformulated objective. Finally, we provide a one-step optimization to fast reach a robust model (Sect. \ref{Sect3.5}). 
\subsection{Preliminaries on Adversarial Training}
\label{Sect3.1}
% Adversarial training is widely accepted as the most effective method in practice to improve the robustness of deep learning models to certain attacks. 
Let us first consider a standard classification task with an underlying data distribution $\mathcal{D}$ over inputs $p\in \mathbb{R}^d$ and corresponding labels $q\in[k]$. The goal then is to find model parameters $\theta$ that minimize the risk $\mathbb{E}_{(p,q)\sim\mathcal{D}}[L(\theta, p, q)]$, where $L(\theta, p, q)$ is a suitable loss function. To improve the model robustness, we wish no perturbations are possible to fool the network, which gives rise to the following formulation:
\begin{equation}
    \min_\theta \rho(\theta),\quad \text{where} \quad \rho(\theta)=\mathbb{E}_{(p,q)\sim\mathcal{D}}[L(\theta, p+\delta, q)],
    \label{eq1}
\end{equation}
where $p+\delta$ refers to the perturbed samples generated by introducing perturbations $
\delta\in\mathcal{S}$ on input data $p$.
$\mathcal{S}$ refers to the allowed perturbation set. 
Eq. (\ref{eq1}) reflects the basic idea of data augmentations. 
% However, as mentioned in \cite{rebuffi2021data}, data augmentation can only be used to improve model robust to a very limited range of perturbations set $\mathcal{S}$. 
% In our case, . 

In contrast, adversarial training improves model robustness more efficiently. By the in-depth study of the landscape of adversarial samples, \cite{madry2017towards} finds the concentration phenomenon of different adversarial samples, which suggests that training on the most aggressive adversary yields
robustness against all other concentrated adversaries.  This gives rise to the formulation of adversarial training which is a saddle point problem:
\begin{equation}
    \min_\theta \rho(\theta),\quad \text{where} \quad \rho(\theta)=\mathbb{E}_{(p,q)\sim\mathcal{D}}[\max_{\delta\in\mathcal{S}}L(\theta, p+\delta, q)].
    \label{eq2}
\end{equation}
The saddle point problem can be viewed as the composition of an inner maximization problem and an outer minimization problem, where the inner maximization problem is finding the worst-case samples for the given model, and the outer minimization problem is to train a model robust to adversarial samples. Compared with data augmentation, adversarial training searches for the best solution to the worst-case optimum and can improve the model robustness to perturbations in larger ranges \cite{madry2017towards}.

% \cite{madry2017towards} show that the adversarial samples found by attack all have similar loss values, both for
% normally trained networks and adversarially trained networks. This concentration phenomenon
% suggests an intriguing view on the problem in which robustness against the most aggressive adversary yields
% robustness against all other adversaries. To this end, instead of feeding samples from the distribution $\mathcal{D}$ directly into the loss $L$, we allow the most aggressive adversary to perturb the input first.
% This gives rise to the formulation of adversarial training which is a saddle point problem:
% \begin{equation}
%     \min_\theta \rho(\theta),\quad \text{where} \quad \rho(\theta)=\mathbb{E}_{(p,q)\sim\mathcal{D}}[\max_{\delta\in\mathcal{S}}L(\theta, p+\delta, q)].
% \end{equation}
% The saddle point problem can be viewed as the composition of an inner maximization problem and an outer minimization problem, where the inner maximization problem is finding the worst-case samples for the given model, and the outer minimization problem is to train a model robust to adversarial examples. 
% % Searching for the best solution to the worst-case optimum can 

% is that attaining small adversarial loss $\rho(\theta)$ gives a guarantee that no allowed attack will fool the network. Here, the attacks can be both the black-box attack (randomly generated perturbations) and white-box attack (pertubations found through the model gradient information).
\subsection{Problem Formulation}
\label{Sect3.2}
Our main goal is to improve the robustness of the point cloud classifiers to rotation attacks through the adversarial training framework. We reformulate Eq. (\ref{eq2}) by specifying the perturbation to be the point cloud rotation as follows:
\begin{equation}
    \min_\theta \rho(\theta),\quad \text{where} \quad \rho(\theta)=\mathbb{E}_{(p,q)\sim\mathcal{D}}[\max_{R\in SO(3)}L(\theta, Rp, q)],
    \label{eq3}
\end{equation}
where $p\in\mathbb{R}^{n\times3}$ refers to an input point cloud of size $n$ and $q\in[k]$ is the corresponding class label. $\theta$ is the parameters of point cloud classifiers such as PointNet \cite{qi2017pointnet} or DGCNN \cite{wang2019dynamic}. $Rp$ refers to the adversarial samples generated by using matrix $R$ to rotate the input $p$ and $SO(3)$ is the group of all rotations around the origin of $\mathbb {R}^{3}$ Euclidean space.
We set the rotation $R\in SO(3)$ to ensure the objective is to make the model robust to arbitrary rotations.

As discussed in \cite{madry2017towards}, one key element for obtaining a good solution to Eq. (\ref{eq3}) is using the strongest possible adversarial samples to train the networks. Following this principle, we first propose a novel rotation attack method that enjoys satisfactory attack success and thus better serves for the adversarial training to improve model robustness. 
% In the following section, we will strictly follow this rule to design complete frameworks to solve the above equations for obtaining classifiers with rotation robustness to arbitrary rotations.

\subsection{Attack---Inner Maximization}
\label{Sect3.3}
For the inner maximization problem, we expect a strong rotation attack algorithm that can find the most aggressive samples inducing high classification loss. A previous study \cite{zhao2020isometry} introduced two rotation attack methods, Thompson Sampling Isometry (TSI) attack and Combined Targeted Restricted Isometry (CTRI) attack, for generating adversarial rotations. However, they can hardly be used in adversarial training for the following reasons: (1) the TSI attack is a black-box attack, which has no direct access to the classifier parameters and thus can hardly be used to find samples inducing high loss. (2) CTRI attack is a white-box attack and one can use parameter information to search the most aggressive samples. Yet, in CTRI, there is no strict constraint for the matrix to be a pure rotation, which leads to adversarial samples with non-rigid deformation. To this end, we propose a novel white-box attack that can efficiently find the most aggressive samples while guaranteeing that the attack is pure rotation.

\textbf{Gradient Descent on Angles.} Firstly, to ensure the attack is pure rotation, we propose to optimize the attack by gradient descent on rotating angles. Specifically, for an n-point cloud $p=[x_i,y_i,z_i], i = 1...n$, we consider vectors $\Phi = [\phi_x, \phi_y, \phi_z]$ with 3 parameters denoting rotation angles along three axes.
% and use $[\frac{\partial L}{\partial \phi_x},\frac{\partial L}{\partial \phi_y},\frac{\partial L}{\partial \phi_z}]$ to denote the gradient back-propagated on rotation angles of three axes. 
Rotating points along $z$ axis by $\delta$ will increase the loss $L$ by $\frac{\partial L}{\partial \phi_z}\delta $, which can then be calculated under the spherical coordinate, by the chain rule as:
\begin{equation}
\begin{aligned}
      \frac{\partial L}{\partial \phi_z} =&\sum_{i=1}^n(\frac{\partial x_i}{\partial \phi_z}\frac{\partial L}{\partial x_i} +\frac{\partial y_i}{\partial \phi_z}\frac{\partial L}{\partial y_i} +\frac{\partial z_i}{\partial \phi_z}\frac{\partial L}{\partial z_i})\\=&  \sum_{i=1}^n(-y_i \frac{\partial L}{\partial x_i}+x_i\frac{\partial L}{\partial y_i}),
      \label{eq4}
\end{aligned}
\end{equation}
where, $\frac{\partial L}{\partial x} =\nabla_{x} L(\theta, p, q)$  and $\frac{\partial L}{\partial y} =\nabla_{y} L(\theta, p, q)$ are gradients back-propagated on point coordinates. For the rest of the rotation axes, $\frac{\partial L}{\partial \phi_x}$ and $\frac{\partial L}{\partial \phi_y}$ can also be calculated in the same way.
Based on Eq. (\ref{eq4}), we can iteratively optimize the angles by gradient descent to obtain adversarial rotations that induce high loss. Finally, the rotation matrix is generated from optimized angles as $R=R_{\phi_z}R_{\phi_y}R_{\phi_x}$, where $R_{\phi_x}$ corresponds to the rotation matrix that rotates $\phi_x$ degrees around $x$ axis. More derivations about the gradient calculation and rotation matrix construction will be provided in the supplementary.

\textbf{Axis-Wise Attack.} In order to efficiently find the most aggressive rotations, based on the angle gradients, we further propose an axis-wise mechanism.  Specifically, we subdivide a rotation in SO(3) into rotations around three axes for optimization. By doing so, each time we can choose the most aggressive axis to rotate, resulting in stronger attacks. We approximate the loss change ratio of a specific axis by $|\frac{\partial L}{\partial \phi}|$, which reflects the influence of rotating around a certain axis on final losses. Next, we select the most influenced axis
\begin{equation}
    \xi^*= \text{argmax}_{\xi}|\frac{\partial L}{\partial \phi_\xi}|, \xi\in[x,y,z],
    \label{eq5}
\end{equation}
and attack the axis by rotating one step in the opposite direction of gradient descent:
\begin{equation}
    \phi^{(t+1)}_{\xi^*} = \phi^{(t)}_{\xi^*} + \alpha\text{sign}(\frac{\partial L}{\partial \phi_{\xi^*}}).
\end{equation}
Compared with simultaneously optimizing on all three axes, the axis-wise attack can specify a gentler change of the rotation angles in each attack step.

% Experiments show that this method results high attack losses.

\textbf{Implementation Details.} In the real implementations, we adopt several other general settings to find adversarial samples. Firstly, we use the Projected Gradient Descent (PGD) \cite{madry2017towards} to optimize angles. Compared with the normal gradient descent, PGD ensures that the optimized angles can be constrained into certain scopes:
\begin{equation}
    \phi^{(t+1)}_{\xi^*} = \text{Proj}_{[-\pi, \pi]}(\phi^{(t)}_{\xi^*} + \alpha\text{sign}(\frac{\partial L}{\partial \phi_{\xi^*}})).
    \label{eq7}
\end{equation}
In our case, we set the projected scope as $[-\pi, \pi]$ to avoid the discontinuity caused by the periodicity of rotation. Then, instead of cross-entropy, we follow \cite{xiang2019generating, zhao2020isometry} to adopt CW loss \cite{carlini2017towards} to modify the cross-entropy as a more powerful adversarial objective to generate stronger adversary.
% \begin{equation}
%     L = \max_{i\neq t}(-\log\mathcal{Z}_i)-(-\log\mathcal{Z}_t),
%     \label{eq8}
% \end{equation}
% where, $\mathcal{Z}_i$ is the i-th element of the predicted probability vector (output of softmax layer), $\mathcal{Z}_t$ refers to the predicted probability of the ground truth class.
% % and $(r)^+$ represents the  $max(r,0)$ operator. 
% Compared with the cross-entropy loss, Eq. (\ref{eq8}) imposes a stronger constraint by simultaneously increasing the predicted probability of the least likely class $\max_{i\neq t}(-\log\mathcal{Z}_i)$ and decreasing the probability of ground truth class $ (-\log\mathcal{Z}_t)$. 
Finally, to make sure that the generated adversary can be more evenly distributed among $[-\pi, \pi]$, we adopt a random start strategy. For each input point cloud, we will initialize it with a random rotation angle, then continue to attack along with the initialization angles. The proposed axis-wise rotation attack algorithm is illustrated in Algorithm (\ref{alg1}).
% Experiments show that our methods benefit from the random start to find more attacks that induces higher loss. 

\begin{algorithm}[t]
\caption{
Axis-Wise Rotation Attack
}
\label{alg1}
\begin{algorithmic}[1]
\REQUIRE Point cloud input $p$, label $q$ and model parameters $\theta$, loss function $L(\theta, p, q)$, number of iterations $T$, step size $\alpha$, initial rotation angles $\Phi = [\phi_x, \phi_y, \phi_z]$ and corresponding rotation matrix $R=R_{\phi_x}R_{\phi_y}R_{\phi_z}$.
% (We use subscripts 1,2,3 to correspond to x,y,z axes.)
\FOR{$t = 0$ to $T$}
\STATE Compute the gradients on coordinates:
\STATE $\frac{\partial L}{\partial p^{(t)}} =[\frac{\partial L}{\partial x^{(t)}},\frac{\partial L}{\partial y^{(t)}},\frac{\partial L}{\partial z^{(t)}}]$.

%  $\frac{\partial L}{\partial x} =\nabla_{x} L(\theta, p^{(t)}, q),$
% \STATE  $\frac{\partial L}{\partial y} =\nabla_{y} L(\theta, p^{(t)}, q), \frac{\partial L}{\partial z} =\nabla_{z} L(\theta, p^{(t)}, q).$

\STATE Compute the gradients on angles by Eq. (\ref{eq4}).

% \STATE Compute the attack direction: \STATE $dx,dy,dz=sign( \frac{\partial L}{\partial \phi_x}), sign(\frac{\partial L}{\partial \phi_y}), sign(\frac{\partial L}{\partial \phi_z}).$
% $\xi^*= \mathop{argmax}_{\xi}|\frac{\partial L}{\partial \phi_\xi}|$
\STATE Determining the target axis by Eq. (\ref{eq5}).
% \STATE $\xi^*= \text{argmax}_{\xi}|\frac{\partial L}{\partial \phi_\xi}|, \xi\in[x,y,z]$ 
\STATE Attack the target axis by Eq. (\ref{eq7}).
% \STATE $\phi^{t+1}_{\xi^*} = \text{Proj}_{[-\pi, \pi]}(\phi^t_{\xi^*} + \alpha\text{sign}(\frac{\partial L}{\partial \phi_{\xi^*}}))$
\STATE Update the rotation matrix:
\STATE $R^{(t+1)} = R_{\phi_x^{(t+1)}}R_{\phi_y^{(t+1)}}R_{\phi_z^{(t+1)}}$
\STATE Obtain the attacked point clouds: $p^{(t+1)} = R^{(t+1)}p$
\ENDFOR

Output $R^{(T)},p^{(T)}$
\end{algorithmic}
\end{algorithm}

\subsection{Defense---Outer Minimization}
\label{Sect3.4}
On the defense side, we use Stochastic Gradient Descent (SGD) \cite{bottou2010large} to re-train the model on the adversarial samples. During experiments, we find that for the original training set $\mathcal{A}$ and its attacked set $\mathcal{B}$ with rotations, directly training on set $\mathcal{B}$ can easily lead to model over-fitting. This behavior is known as label leaking \cite{kurakin2016adversarial} and stems from the fact that the gradient-based attack produces a very restricted set of adversarial examples that the network can overfit. The problem can be even worse on the smaller training set, in our case, ModelNet40 \cite{wu20153d}.
To solve the label leaking caused over-fitting problems, we propose to increase the training data with more kinds of adversarial rotations. 
A simple solution is to construct the training set $\mathcal{B}$ with multiple attack $ \mathcal{B}=[\text{attack}_1(\mathcal{A}),\text{attack}_2(\mathcal{A}),…,\text{attack}_i(\mathcal{A})]$. However, multiple attacks can be very time-consuming. To this end, we construct a rotation pool to increase the diversity of training data in a more efficient manner. 
\begin{figure}
    \centering
    \includegraphics[scale=0.9]{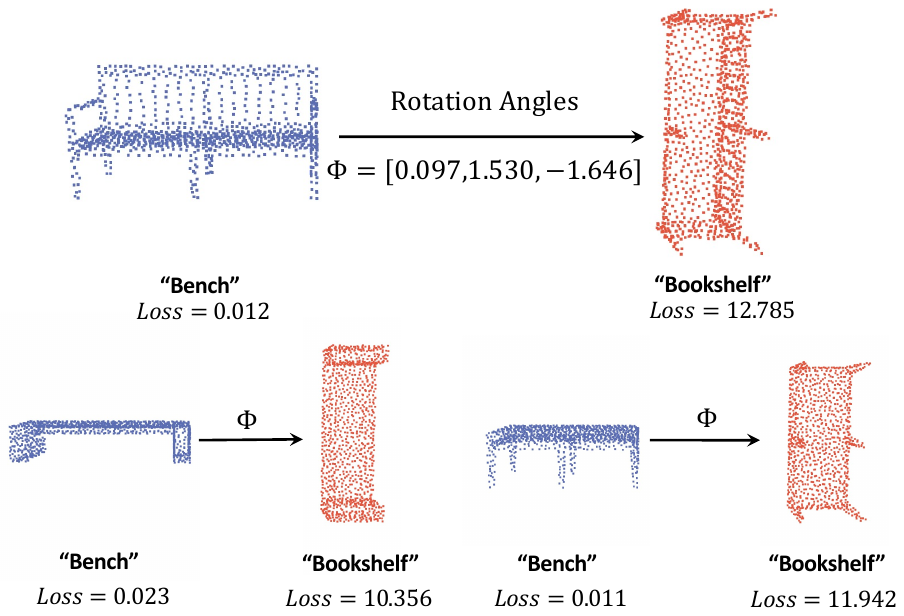}
    \caption{Transferability of adversarial rotations among samples in the same categories. The adversarial rotation found on one sample in “Bench” can be applied to other samples of the same category to induce high loss and mislead the model to classify them into a wrong category “Bookshelf”.}
    \label{fig2}
\end{figure}

\textbf{Rotation Pool.} As shown in Fig. (\ref{fig2}), we observe that the adversarial rotation found on one sample has a strong transferability on other samples of the same category. Based on this observation, instead of saving the rotated samples, we suggest saving the rotation angles produced on each sample by class to construct a rotation pool:
\begin{equation}
    \mathcal{R} = \left[\{\Phi_{i,1} \}_{i=1}^{n_1},\cdots,\{\Phi_{i,k} \}_{i=1}^{n_k},\cdots,\{\Phi_{i,K} \}_{i=1}^{n_K}\right],
    \label{eq9}
\end{equation}
where $\Phi_{i,k}$ is the rotation found on sample $i$ of category $k$. We will save the rotations corresponding to all $n_k$ samples in the category $k$ and traverse all $K$ categories to construct the final rotation pool $\mathcal{R}$.
During defense training, we only need to sample rotations from the rotation pool according to the category to transform the input into adversaries. Thanks to the transferability, the adversarial samples generated by the rotation pool can also induce high classification loss. Experiments in Sect. \ref{sect4.5} also confirm that the rotation pool can effectively solve the over-fitting problem.
\begin{figure*}
    \centering
    \includegraphics[scale=1.01]{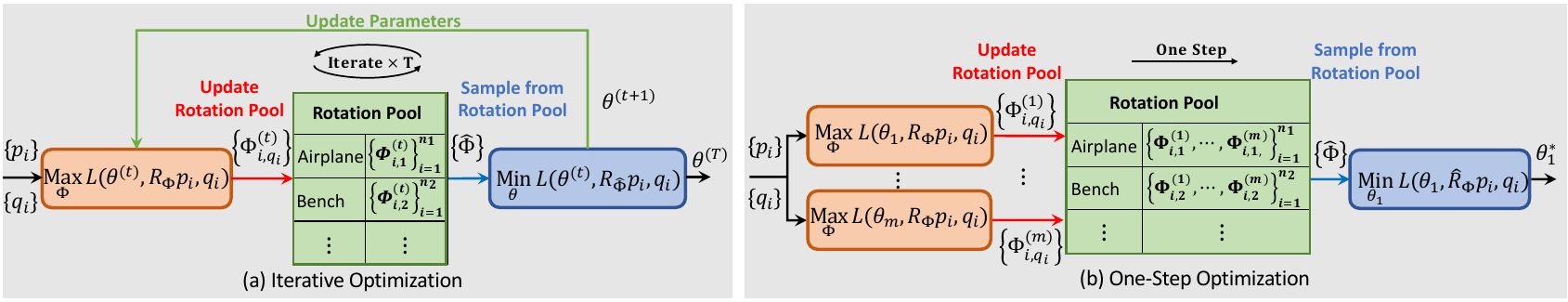}
    \caption{Comparison of different optimizations.
    For the iterative optimization (a), model with parameters $\theta$ will be repeatedly optimized on the min-max problem $T$ times until converging to a robust parameter $\theta^{T}$. In contrast, the proposed one-step optimization (b) constructs the rotation pool by attacking $m$ different models and requires only one step to obtain robust parameters of the targeted model.}
    \label{fig3}
\end{figure*}

\textbf{Iterative Optimization.} In order to solve the minimization problem, \emph{i.e.} Eq. (\ref{eq3}), in adversarial training to reach the final robust models, an iterative optimization scheme is usually adopted. Specifically, in the first iteration, we will attack the pre-trained classifier to initialize the rotation pool and then re-train the classifier on adversarial samples generated from the rotation pool towards a robust model. In the following iterations, we will attack the latest robust model to update the rotation pool iteratively:
\begin{equation}
    \Phi^{(t)}_{i,q_i} =\max_{\Phi} L(\theta^{(t)}, R_\Phi p_i, q_i),
\end{equation}
where $\theta^{(t)}$ refers to the parameters of robust model after $t$ iterations, $R_\phi$ is the rotation matrix of random start angles $\phi$ and $q_i$ is the class label corresponding to input sample $p_i$. $\Phi^{(t)}_{i,q_i}$ refers to the rotation found on sample $i$ of category $q_i$ in the $t$-th iteration. 
We then re-train the classifier on the adversaries generated from the updated pool $\mathcal{R}^{(t)}$ to reach a more robust model. 
The process will be repeated until the model converges to the most robust state.
% The most important advantage of such iterative optimization is that by progressively enhancing model robustness and updating the rotation pool, we can obtain more aggressive rotations that can never be produced in one step iteration. 
% We visualize the effectiveness of iterative optimization on progressively improving model rotation robustness in Sect. 4.5.
% During experiments, in one iteration, we use 10 epochs to attack pre-trained model to construct rotation pools and another 100 epochs to re-train the model on adversarial samples generated from rotation pool. We adopt a total of ten iterations, to progressively reach the final robust model. We illustrate the iterative process in Fig. \ref{fig2}.and visualizing its ability on progressively improving model rotation robustness in Sect. 4.5.

\subsection{One-Step Optimization}
\label{Sect3.5}
The naive implementation above requires multiple iterations on both the attack and defense sides. Though obtaining robust models, the whole process is extremely time-consuming. Inspired by the ensemble adversarial training (EAT) \cite{tramer2017ensemble}, we further propose an efficient one-step optimization to reach the robust model with lower training cost. 

Specifically, instead of iterating multiple times for obtaining more aggressive samples, EAT proposes to introduce the adversarial examples crafted on other stronger static pre-trained models. Intuitively, as adversarial samples transfer between models, perturbations crafted on the more robust model are good approximations for the maximization problem of the target model. We follow this principle to solve the minimization problem Eq. (\ref{eq3}) in one step. Concretely, we not only attack the target classifier but attack more robust classifiers to construct a larger rotation pool:
\begin{equation}
    \Phi^{(m)}_{i,q_i} =\max_{\Phi} L(\theta_m, R_\Phi p_i, q_i),
    \label{eq11}
\end{equation}
where, $\theta_m$ refers to the parameters of model $m$ and $\Phi^{(m)}_{i,q_i}$ is the adversarial rotation generated by attacking model $m$. By attacking $m$ models, the resulting rotation pool has $m$ times more aggressive rotations than the iterative optimization does.
For defense, similar to the iterative optimization, we use the adversarial rotation sampled from the rotation pool to re-train the target model.
Compared with the iterative manner, the one-step optimization achieves competitive results with faster training progress. Hence, we select the one-step optimization as the default implementation of our ART-Point framework. The comparison between the two optimization methods is shown in Fig. (\ref{fig3}). Detailed implementations and comparison experiments will be provided in the supplementary.

\section{Experiments}
% We conduct extensive experiments to verify the effectiveness of the proposed ART-Point for improving the robustness of point cloud classifiers to rotations.
\subsection{Experiment Setup}
\label{sect4.1}
\textbf{Datasets.} We evaluate our methods on two classification datasets ModelNet40 \cite{wu20153d} and ShapeNet16 \cite{yi2016scalable}. ModelNet40 contains 12,311 meshed CAD models from 40 categories. ShapeNet16 is a larger dataset which contains 16,881 shapes from 16 categories.
For both datasets, we follow the official train and test split scheme and use the same data pre-processing as in \cite{qi2017pointnet,qi2017pointnet++,wang2019dynamic} where each model is uniformly sampled with 1,024 points from the mesh faces and rescaled to fit into the unit sphere. 

\textbf{Models.} We select three point cloud classifiers to evaluate our method, including PointNet \cite{qi2017pointnet}, a pioneer network that processes points individually, PointNet++ \cite{qi2017pointnet++}, a hierarchical feature extraction network and DGCNN \cite{wang2019dynamic}, a graph-based feature extraction
network. These classifiers lack robustness to rotation. By verifying these classifiers, we show that ART-Point can be applied to various learning architectures to improve rotation robustness.

% \textbf{Implementation Details.}
% We implement ART-Point using PyTorch \cite{paszke2017automatic}. In detail, during attacking, we set the step size of angle descent $
% \alpha=0.01$, a batch size $B=17$ and adopt ten steps descent to obtain the final adversarial rotations. During defensing, we mainly use SGD to train different classifiers adopting the same optimizer and learning rate schedules as proposed in their papers.

\textbf{Evaluations.} 
In order to comprehensively compare the rotation robustness of different models, we design three evaluation protocols:
(1) Attack. The test set is adversarially rotated by the proposed attack algorithm for evaluating model defense.
(2) Random. The test set is 
randomly rotated for evaluating model rotation robustness. 
(3) Clean. The test set is unchanged for evaluating the discriminative ability under aligned data. 
Moreover, we use the attack success rate to evaluate our attack algorithm. The attack success rate is calculated as the percentage of correctly predicted samples in the test set before and after the attack.
\begin{table}[t]
    \centering
    \renewcommand\arraystretch{1.1}
    \setlength{\tabcolsep}{0.5mm}{
    \begin{tabular}{l|ccc}
    \hline
    \multirow{2}{*}{Method}   & \multicolumn{3}{c}{ModelNet40}  \\\cline{2-4}
    &Attack&Random&Clean\\\hline
    PointNet \cite{qi2017pointnet} (RA) & 55.6 & 74.4 & 76.7\\
    PointNet++ \cite{qi2017pointnet++} (RA) & 58.9 &80.1&82.3\\
    DGCNN \cite{wang2019dynamic} (RA)  &  65.6 &  85.7 & 87.6 \\\hline
    ART-PointNet (\textbf{Ours}) & 85.6(30.0$\uparrow$)  &84.3(9.9$\uparrow$)  & 85.5(8.8$\uparrow$) \\
    ART-PointNet++ (\textbf{Ours})& 90.1(31.2$\uparrow$)    & 87.5(7.4$\uparrow$)   & 88.6(6.3$\uparrow$)   \\
    ART-DGCNN (\textbf{Ours}) & \textbf{91.5}(25.9$\uparrow$) & \textbf{90.5}(4.8$\uparrow$)  & \textbf{91.3}(3.7$\uparrow$)   \\\hline\hline
    \multirow{2}{*}{Method}   & \multicolumn{3}{c}{ShapeNet16}  \\\cline{2-4}
    &Attack&Random&Clean\\\hline
    PointNet \cite{qi2017pointnet} (RA) &66.4  & 87.3   &89.5  \\
    PointNet++ \cite{qi2017pointnet++} (RA)&70.5 &89.7  &92.1 \\
    DGCNN \cite{wang2019dynamic} (RA)  &74.4 & 90.5     &94.3 \\\hline
    ART-PointNet (\textbf{Ours})  &96.9(30.5$\uparrow$)   &95.1(7.8$\uparrow$)  & 96.2(6.7$\uparrow$)  \\
    ART-PointNet++ (\textbf{Ours}) &97.8(27.3$\uparrow$)  &96.3(6.6$\uparrow$) & 97.5(5.4$\uparrow$)  \\
    ART-DGCNN (\textbf{Ours})      &\textbf{98.4}(24.0$\uparrow$)   &\textbf{97.7}(7.2$\uparrow$)  & \textbf{98.1}(3.8$\uparrow$)  \\\hline

    \end{tabular}}
    \caption{Comparing three evaluation protocols under ModelNet40 \cite{wu20153d} and ShapeNet16 \cite{yi2016scalable} for classifiers trained via rotation augmentation (RA) and adversarial rotation (ART).}
    \label{tab1}
\end{table}
\subsection{Comparison with Rotation Augmentation}
We first compare the effectiveness of the proposed ART-Point with rotation augmentation (RA) for improving model rotation robustness. For classifiers using rotation augmentation, we will train them with randomly rotated inputs. In Tab. (\ref{tab1}), we illustrate the comparison results under ModelNet40 \cite{wu20153d} and ShapeNet16 \cite{yi2016scalable}. 
From the table, several observations can be obtained. 
Firstly, compared with rotation augmentation, the proposed ART-Point results in models performing better under all protocols. Such performance improvements can be consistently observed on all three classifiers under both datasets. Secondly, under the attacked test set, the classification accuracy of model trained using ART-point is significantly higher than model trained with RA. (maximum increase: 31.2\%). This is mainly because that rotation augmentation can hardly defend against adversarial rotations found using model gradient information. In contrast, our method shows stronger defense to adversarial rotations. We will further test the defense ability of our method under different rotation attacks in Sect. \ref{sect4.4}. Both observations suggest that the proposed ART-Point is a more effective method to improve the rotation robustness of point cloud classifiers than rotation augmentation.

% models trained with ART-Point  the attacked test set far exceeds random augmentation.
% Secondly, the performance improvement of our method on the attacked test set is more obvious

% proposed ART-Point training scheme  

% . Firstly, the performance of all classifiers significantly decrease under rotated inputs. Simple random rotation can greatly reduce the accuracy of recognition. Using the rotating attack, we designed, the performance of the model becomes worse.
% Secondly, after using AT, classifiers can be better defensed to our rotation attack. The random rotation which can be regarded as a weak attack is also defensed by our methods. It demonstrates the effectiveness of our method for improving model robustness to random rotations. Finally, our method will not damage the recognition ability of the model on original datasets too much, that is, the performance will not drop too much on the pre-aligned clean dataset. 

\subsection{Comparison with Rotation Robust Classifiers}
We further compare robust models trained by ART-Point with existing rotation robust classifiers, including \cite{rao2019spherical,zhang2019rotation,chen2019clusternet,li2021rotation} that convert point clouds into rotation invariant descriptors and \cite{thomas2018tensor,shen20203d,deng2021vector,chen2021equivariant} that design rotation-equivariant architectures, to further illustrate appealing properties of our method.  Rotation robust classifiers will be trained on random rotated inputs. The comparison results based on all protocols under ModelNet40 \cite{wu20153d} are shown in Tab. (\ref{tab2}).
\begin{table}[t]
    \centering
    \renewcommand\arraystretch{1   }
    \setlength{\tabcolsep}{3mm}{
    \begin{tabular}{l|ccc}
    \hline
    \multirow{2}{*}{Method}   & \multicolumn{3}{c}{ModelNet40}  \\\cline{2-4}
    &Attack&Random&Clean\\\hline
    \multicolumn{4}{l}{\textit{Classifiers Using Invariant Descriptors}}\\\hline    
    SFCNN \cite{rao2019spherical} &90.1 & 90.1 &90.1 \\
    RI-Conv \cite{zhang2019rotation} &86.5 &86.4 &86.5\\
    ClusterNet \cite{chen2019clusternet}&87.1  &87.1 &87.1 \\
    RI-Framework\cite{li2021rotation}&89.4 & 89.3 &89.4 \\\hline
    \multicolumn{4}{l}{\textit{Classifiers with Equivariant Architectures}}\\\hline
    TFN \cite{thomas2018tensor}&87.6 &87.6&87.6 \\
    REQNN \cite{shen20203d}&74.4 &74.1&74.4\\
    VN-PointNet \cite{deng2021vector}&77.2&77.2&77.2\\
    VN-DGCNN\cite{deng2021vector}&90.2&90.2&90.2\\
    EPN \cite{chen2021equivariant}&88.3 &88.3&88.3\\\hline
    \multicolumn{4}{l}{\textit{Ours}}\\\hline 
    ART-PointNet  & 85.6  &84.3  & 85.5 \\
    ART-PointNet++ & 90.1    & 87.5  & 88.6   \\
    ART-DGCNN & \textbf{91.5} & \textbf{90.5}  & \textbf{91.3}  \\\hline
    \end{tabular}}
    \caption{Comparing three evaluation protocols under ModelNet40 \cite{wu20153d} for various rotation robust classifiers.}
    \label{tab2}
\end{table}
% Note that while both the proposed ART-Point and other robust methods need to train the model on rotated samples, our method can train via the existing classifiers without architectural modifications. 
Firstly, our best model ART-DGCNN outperforms all equivariant or invariant methods under three evaluation protocols, which indicates its stronger robustness over rotations. 
Secondly, both equivariant or invariant methods perform similarly under all protocols, which is undesirable, since the clean test set should more easily be classified by the model. This is mainly because that these methods obtain rotation robustness by separating the pose information from point clouds via modifications on input space or model architectures.
In contrast, ART-Point uses original classifiers for training on adversarial samples in 3D space, the resulting model 
not only better inherits the performance of original classifiers on clean sets but shows great defense on the attacked test set.

% \textbf{try to separate the pose information from point clouds via modifications on input space or model architectures for obtaining rotation robustness which results in}

\subsection{Attack and Defense}
\label{sect4.4}
Beyond rotation robustness, our method provides a complete set of tools for attack and defense on point cloud classifiers. To verify the proposed attack algorithm, we compare the attack success rate of our method with other rotation attacks proposed in \cite{zhao2020isometry}.
Meanwhile, we also show the defense ability of classifiers trained with ART-Point. The results are illustrated in Tab. (\ref{tab3}).
\begin{table}[t]
    \centering
    \renewcommand\arraystretch{0.9}
    \setlength{\tabcolsep}{1mm}{
    \begin{tabular}{l|ccc}
    \hline
    \multirow{2}{*}{Models}& \multicolumn{3}{c}{Rotation Attack Algorithm}\\ \cline{2-4}
          &TSI \cite{zhao2020isometry}&CTRI \cite{zhao2020isometry}&Ours\\\hline
    PointNet \cite{qi2017pointnet} & 96.92   & 99.44&99.54\\
    PointNet++ \cite{qi2017pointnet++} &91.31 &97.93&98.96\\
    DGCNN \cite{wang2019dynamic}  &89.81 &97.99& 98.51 \\\hline
    ART-PointNet (\textbf{Ours})  &9.71  &11.13
    &12.78\\
    ART-PointNet++ (\textbf{Ours})   &4.31 &6.60
    &7.92 \\
    ART-DGCNN (\textbf{Ours}) &3.14  &5.33 &6.62 \\\hline
    
    \end{tabular}}
    \caption{Comparing attack success rate  (\%) of several attack algorithms on different classifiers under ModelNet40 \cite{wu20153d}.}
    \label{tab3}
\end{table}
In the first three rows, we report the attack success rate of different attack algorithms on classifiers trained using clean samples. As can be seen, compared with the other two rotation attacks, our attack achieves the highest success rate on all three classifiers. 
In the last three rows, we further report the attack success rate on classifiers trained using ART-Point. As can be seen, ART-Point improves model defense against rotation attacks.

% It can be seen that our method not only guarantees the attack success rate, but also guarantees a lower loss value, thereby meeting the requirements of adversary training.
% We further illustrate the loss changes with our attack iterating in Fig3. 
% In the second row of the table, we illustrate the defensiveness of our method against different attacks. It can be seen that our models trained with stronger attack samples naturally has better defense against both attacks proposed in \cite{zhao2020isometry}, which is in line with the law of adversarial training.

\subsection{Ablation Study}
\label{sect4.5}
Finally, we conduct ablation studies to prove the effectiveness of our designs in ART-Point. 
All ablation experiments are conducted on the PointNet \cite{qi2017pointnet} classifier and evaluated under randomly rotated test sets\footnote{More ablation studies on descent step, rotation angle, and attack step size can be found in the supplementary material.}.

\textbf{Different Attacks.} We use adversarial samples generated by different rotation attacks for adversarial training and investigate the impact on the robustness of the resulting models. We adopt several attacks to generate adversarial samples that induce different loss values, including the random rotation attack, attacks in \cite{zhao2020isometry} and our attacks with different steps. 
In the left column of Tab. (\ref{tab4}), we illustrate the average classification loss of samples produced by different attacks and results of adversarial training using corresponding samples. Compared with other attacks, the proposed axis-wise rotation attack with 10 steps gradient descent induces the highest loss value. 

% Moreover, the experimental results prove that using aggressive samples with higher loss leads to better adversarial training results.
\begin{table}[t]
    \centering
    \renewcommand\arraystretch{0.9}
    \setlength{\tabcolsep}{0.8mm}{
    \begin{tabular}{l|cc||l|cc}
    \hline
    Methods&Loss&Acc.&Methods&Loss&Acc.\\\hline   Random &5.13 &74.4& w/o RP&12.72 & 55.8\\
    TSI \cite{zhao2020isometry}&7.35 &79.5&RP(pn1) &10.19 & 82.9\\
    CTRI\cite{zhao2020isometry}&8.87&82.1  &  RP(pn1,pn2) &12.01&82.6\\
    Ours (step=1)&7.65 &81.5 & RP(pn1,dg)&12.55& 83.1\\
    Ours (step=5)&9.57 &82.8 & RP(pn2,dg)&13.03& 84.0\\
    Ours (step=10)&\textbf{13.49}&\textbf{84.3} &RP(pn1,pn2,dg)&\textbf{13.49}&\textbf{84.3}\\\hline
    
    \end{tabular}}
    \caption{The average loss of adversarial samples generated by different methods and accuracy of corresponding adversarial training.
    RP(pn1) refers to the rotation pool generated by attacking PointNet \cite{qi2017pointnet}. pn2 and dg refer to PointNet++ \cite{qi2017pointnet++} and DGCNN \cite{wang2019dynamic}.}
    \label{tab4}
\end{table}

\textbf{Rotation Pool.} We verify the necessity of constructing the rotation pool. We compare the results of adversarial training with and without rotation pools. Moreover, we also investigate the impacts of constructing rotation pools from different models.
As shown in the right column of Tab. (\ref{tab4}), although adversarial training without rotation pool generates samples inducing high loss values, the final result is worse than training with rotation pool due to the over-fitting caused by label leaking \cite{kurakin2016adversarial}.

\textbf{Axis-Wise Attack.}
We compare our proposed axis-wise rotation attack with the standard attack algorithm, which simultaneously optimizes three angles in one gradient descent. We mainly follow \cite{madry2017towards} to show the average loss value of attacked samples in each step.
% which can effectively reflect the efficiency and strength of the attack algorithm. 
We restart the attack 20 times with random angle initialization. The comparison results are shown in Fig. (\ref{fig2}). As can be seen, the axis-wise mechanism enables the attack algorithm to find more aggressive rotated samples.
% (high value of classification loss).

% As can be seen, the axis-wise mechanism enables the attack algorithm to find more aggressive rotated samples (high value of classification loss). More importantly, we can observe that the axis-wise attack enjoys fewer loss oscillations in the initial steps, which makes it possible to find the maximum value faster than the standard attack. Finally, the loss concentration phenomenon \cite{madry2017towards} of adversarial training is also observed, where all attacked samples eventually converge to a stable average loss value despite under different angles of initialization. This phenomenon also demonstrates that using adversarial training to solve the problem of rotation robustness in point cloud classifiers is theoretically feasible.

\begin{figure}[t]
  \centering
  \includegraphics[width=0.8\linewidth]{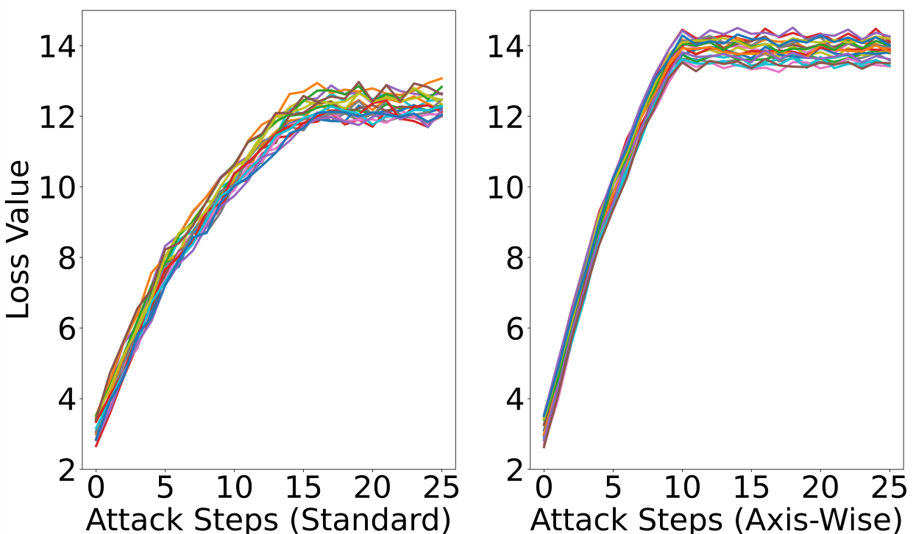}
  \caption{Averaged loss values of attacked samples produced by standard attack and axis-wise attack under different attack steps.}
%   Each plot shows how the loss curve evolves during 20 runs of attacks where each run starts at a uniformly random rotation angle.}
  \label{fig2}
\end{figure}

\subsection{Discussions of Limitations and Society Impact}
Since our method is mainly based on adversarial training, one limitation is that we need to obtain a fully trained model with accessible parameters in the first place. Meanwhile, since our method involves a rotating attack algorithm, it may be exploited for attacking point cloud based 3D object detection systems, which is a potential negative societal impact.
\section{Conclusion}
In this paper, we propose ART-Point to improve the rotation robustness of point cloud classifiers via adversarial training. ART-Point consists of an axis-wise rotation attack and a defense method with the rotation pool mechanism. It can be adopted on most existing classifiers with fast one-step optimization to obtain rotation robust models. Experiments show that the novel rotation attack achieves a high attack success rate on most point cloud classifiers. Moreover, our best model ART-DGCNN shows great robustness to arbitrary and adversarial rotations and outperforms existing state-of-the-art rotation robust classifiers.

% We propose an axis-wise rotation attack algorithm to find the most adversarial samples and design optimization scheme to retrain the model based on these adversaries. 

% Experiments show that our method can effectively improve model's rotation robustness. Compared with existing rotation robust classifiers, we achieve better results on both rotated and clean datasets. Moreover, adversarial training allows the model to be defensive against rotation attacks. Finally, extensive ablation studies prove the effectiveness and necessity of each module design.

%%%%%%%%% REFERENCES
{\small
\bibliographystyle{ieee_fullname}
\bibliography{egbib}
}

\newpage
\appendix
\section{Overview}
This document provides technical details, additional quantitative results, and more qualitative test examples to the main paper.
In Sect. \ref{s2} we provide derivations about the gradients back-propagated on three rotation angles and illustrate the construction of rotation matrices. 
In Sect. \ref{s3} we show more implementation details on our network architectures and training parameters. Then Sect. \ref{s4} illustrates comparison experiments between different optimization skills, while Sect. \ref{s5} shows more analysis experiments on our attack algorithm. At last, we show some visualization results in Sect. \ref{s6}.

%-------------------------------------------------------------------------
\section{Gradient Derivation and Rotation matrix Construction (Sect. 3.3)}
\label{s2}

\textbf{Gradient Derivation.}
As illustrated in the main paper, rotating points along $z$ axis by $\delta$ will increase the loss $L$ by $\frac{\partial L}{\partial \phi_z}\delta $, where $\frac{\partial L}{\partial \phi_z}$ can be calculated by the chain rule as: 

\begin{equation}
\begin{aligned}
      \frac{\partial L}{\partial \phi_z} =&\sum_{i=1}^n(\frac{\partial x_i}{\partial \phi_z}\frac{\partial L}{\partial x_i} +\frac{\partial y_i}{\partial \phi_z}\frac{\partial L}{\partial y_i} +\frac{\partial z_i}{\partial \phi_z}\frac{\partial L}{\partial z_i}).
    %   \\=&  \sum_{i=1}^n(-y_i \frac{\partial L}{\partial x_i}+x_i\frac{\partial L}{\partial y_i}),
      \label{e1}
\end{aligned}
\end{equation}
Here, $\phi_z$ refers to the rotation angle around $z$ axis, which is the same as the azimuthal angle $\phi$ in the following spherical coordinate system $(r,\theta, \phi)$:
\begin{equation}
\left\{
        \begin{aligned}
             &x=r\cos\phi\sin\theta,   \\
             &y=r\sin\phi\sin\theta , \\
             &z=r\cos\theta.
        \end{aligned}
\right.
\label{e2}
\end{equation}
Then, based on Eq. (\ref{e2}), we can write Eq. (\ref{e1}) as follows:
\begin{equation}
\begin{aligned}
      \frac{\partial L}{\partial \phi_z} =&\sum_{i=1}^n(\frac{\partial x_i}{\partial \phi}\frac{\partial L}{\partial x_i} +\frac{\partial y_i}{\partial \phi}\frac{\partial L}{\partial y_i} +\frac{\partial z_i}{\partial \phi}\frac{\partial L}{\partial z_i})\\
      =& \sum_{i=1}^n(-r\sin\phi\sin\theta*\frac{\partial L}{\partial x_i} +r\cos\phi\sin\theta*\frac{\partial L}{\partial y_i})
      \\=&  \sum_{i=1}^n(-y_i \frac{\partial L}{\partial x_i}+x_i\frac{\partial L}{\partial y_i}).
      \label{e3}
\end{aligned}
\end{equation}

Similarly, for the remaining rotation axes $\phi_x$ and $\phi_y$, we can calculate the gradients simply by rolling the coordinate system in Eq. (\ref{e3}) as follows:
\begin{equation}
\begin{aligned}
      \frac{\partial L}{\partial \phi_x}
      =&  \sum_{i=1}^n(-z_i \frac{\partial L}{\partial y_i}+y_i\frac{\partial L}{\partial z_i}), \\
      \frac{\partial L}{\partial \phi_y}
      =&  \sum_{i=1}^n(-x_i \frac{\partial L}{\partial z_i}+z_i\frac{\partial L}{\partial x_i}).
      \label{e4}
\end{aligned}
\end{equation}

\textbf{Rotation Matrix Construction.}
Given the optimized rotation angle $\Phi=[\phi_x, \phi_y, \phi_z]$, we construct the corresponding rotation matrices as follows:
\begin{equation}
R_{\phi_x} = \left[ 
\begin{array}{ccc}
1 & 0 & 0\\
0 & \cos\phi_x & -\sin\phi_x \\
0 & \sin\phi_x & \cos\phi_x
\end{array} 
\right],
\label{e5}
\end{equation}

\begin{equation}
R_{\phi_y} = \left[ 
\begin{array}{ccc}
\cos\phi_y & 0 & \sin\phi_y\\
0 & 1 & 0 \\
-\sin\phi_y & 0 & \cos\phi_y
\end{array} 
\right],
\label{e6}
\end{equation}

\begin{equation}
R_{\phi_z} = \left[ 
\begin{array}{ccc}
\cos\phi_z & -\sin\phi_z & 0\\
\sin\phi_z & \cos\phi_z & 0  \\
0 & 0 & 1
\end{array} 
\right].
\label{e7}
\end{equation}
Based on above equations, we compute the final rotation matrix $R=R_{\phi_z}\cdot R_{\phi_y}\cdot R_{\phi_x}$, where ``$\cdot$'' refers to the matrix multiplication. 

\section{Implementation Details}
\label{s3}
We implement ART-Point using PyTorch \cite{paszke2017automatic}. In detail, during attack, we set the step size of angle gradient descent $
\alpha=0.01$, a batch size $B=17$ and adopt ten steps descent to obtain the adversarial rotation. During defense, we mainly use SGD to train existing point cloud classifiers following the same optimizer and learning rate schedules as used in their papers. We experiment with two optimization methods: iterative optimization and one-step optimization.

For the iterative optimization, we alternate the min-max process until the model converges. Specifically, to train a robust PointNet, in each iteration we use 10 epochs gradient descent on angles for maximization to find the most aggressive rotation angles and 50 epochs for minimization to train on adversarial datasets. We perform 10 iterations in total to obtain the final robust model. 

For the one-step optimization, we construct the rotation pool by attacking multiple classifiers and reach the robust model in a single min-max iteration. Concretely, suppose that our target model is the PointNet classifier \cite{qi2017pointnet}. We not only attack PointNet but attack more robust classifiers such as PointNet++ \cite{qi2017pointnet++} and DGCNN \cite{wang2019dynamic} to construct the rotation pool. We use 10 epochs gradient descent for maximization to find adversarial samples and 200 epochs for minimization to train on adversarial samples.
\begin{figure}[t]
  \centering
  \includegraphics[width=0.75\linewidth]{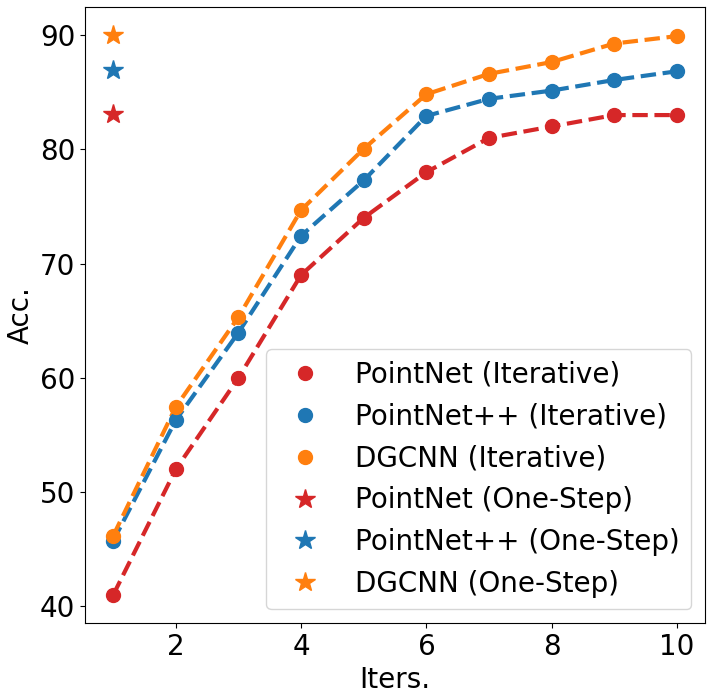}
  \caption{Adversarial training results of three classifiers under ModelNet40 \cite{wu20153d} with different optimizations.}
  \label{fig1}
\end{figure}
\section{Comparison of Different Optimizations}
\label{s4}
We compare the training progress of the naive iterative optimization with the proposed one-step optimization.
The experiments are conducted under ModelNet40 \cite{wu20153d} and resulting classifiers are tested under randomly rotated datasets for evaluating the rotation robustness. We record the performance of three classifiers in each iteration and compare the final results with classifiers trained via the one-step method. 

Specifically, we follow the detailed implementations for both optimizations in Sect. \ref{s3} to reach robust models. It can be seen from Fig. (\ref{fig1}) that for the iterative optimization it usually takes 8-10 iterations to reach the most robust model. In contrast, the one-step method obtains the robust model with competitive performance in one iteration. Note that, for different classifiers in one-step optimizations, the rotating pools are all constructed by attacking three models, \emph{i.e.} PointNet \cite{qi2017pointnet}, PointNet++ \cite{qi2017pointnet++} and DGCNN \cite{wang2019dynamic}.

\section{More Ablation Studies}
\label{s5}
Here, we provide more control experiments to verify our rotation attack algorithm. We mainly conduct studies based on ModelNet40 \cite{wu20153d} with PointNet classifiers \cite{qi2017pointnet}.
% We conduct extensive experiments to verify the effectiveness of the proposed ART-Point for improving the robustness of point cloud classifiers to rotations.

\textbf{Attack Step Size.}
We further illustrate experiments to select the appropriate step size in angle attacks. 
The results are shown in Tab. (\ref{tab1}), where we record the average loss value of attacked samples under different step size $\alpha\ (\text{rad})$. Our attack algorithm finds the most aggressive attacked samples that induce the highest loss with $\alpha=0.01$.  

\textbf{Descent Steps and Rotation Angles.} Finally, we verify the effect of different hyper-parameters on adversarial training. 
We adopt different descent steps during attacking and we also study the performance of our method under limited rotation ranges. The final results are shown in Tab. (\ref{tab5}).
\begin{table}[t]
    \centering
    \renewcommand\arraystretch{1}
    \setlength{\tabcolsep}{1mm}{
    \begin{tabular}{|l|c|c|c|c|}
    \hline
    Descent  & $s=9$& $s=10$& $s=11$& $s=12$ \\ \cline{2-5}
    Steps&83.9&\textbf{84.3}&84.3&84.2 \\\hline
    
    Rotation&$[-\frac{1}{4}\pi, \frac{1}{4}\pi]$&$[-\frac{1}{2}\pi, \frac{1}{2}\pi]$&$[-\frac{3}{4}\pi, \frac{3}{4}\pi]$&$[-\pi, \pi]$\\\cline{2-5}
    Angles&\textbf{87.2} &86.4 &85.5&84.3\\\hline
    \end{tabular}}
    \caption{Adversarial training results under different settings.}
    \label{tab5}
\end{table}

The adversarial training results tend to be saturated when the gradient descent step is large than 10, so we set the attack algorithm with 10 steps descent by defaults. Our method obtains better results under smaller rotation ranges, which demonstrates that by specifying the range of rotation angles, ART-Point can further increase the model robustness. 

\begin{table}[t]
    \centering
    \renewcommand\arraystretch{1}
    \setlength{\tabcolsep}{0.5mm}{
    \begin{tabular}{ccccc}
    \hline
    $\alpha=0.1$& $\alpha=0.08$& $\alpha=0.06$& $\alpha=0.04$&$\alpha=0.02$ \\
   5.3&7.4&9.5&8.9&11.3 \\\hline
         $\alpha=0.01$& $\alpha=0.008$& $\alpha=0.006$& $\alpha=0.004$& $\alpha=0.002$ \\ 
        \textbf{13.5}&12.4&11.7&10.2&9.5 \\\hline
    \end{tabular}}
    \caption{Averaged loss values of attacked samples produced by attacks with different step sizes.}
    \label{tab1}
\end{table}

\begin{figure*}[h]
  \centering
  \includegraphics[width=1\linewidth]{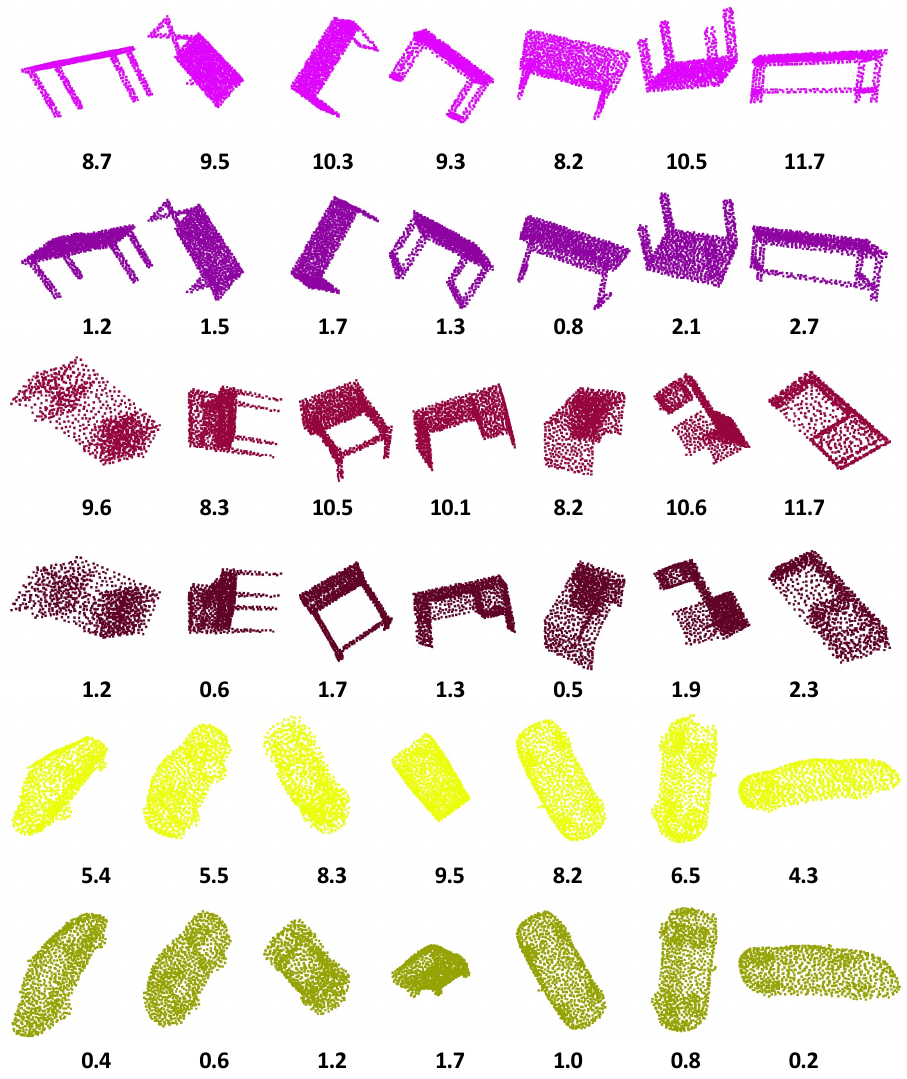}
  \caption{In every two rows, we compare the classification loss of DGCNN \cite{wang2019dynamic} (top row) and ART-DGCNN (bottom row) on the same arbitrarily rotated point clouds, which are randomly sampled from test sets of ModelNet40 \cite{wu20153d}. From top to bottom, the categories of point clouds are ``table'', ``desk'' and ``car''.}
  \label{fig3}
\end{figure*}

\begin{figure*}[h]
  \centering
  \includegraphics[width=1\linewidth]{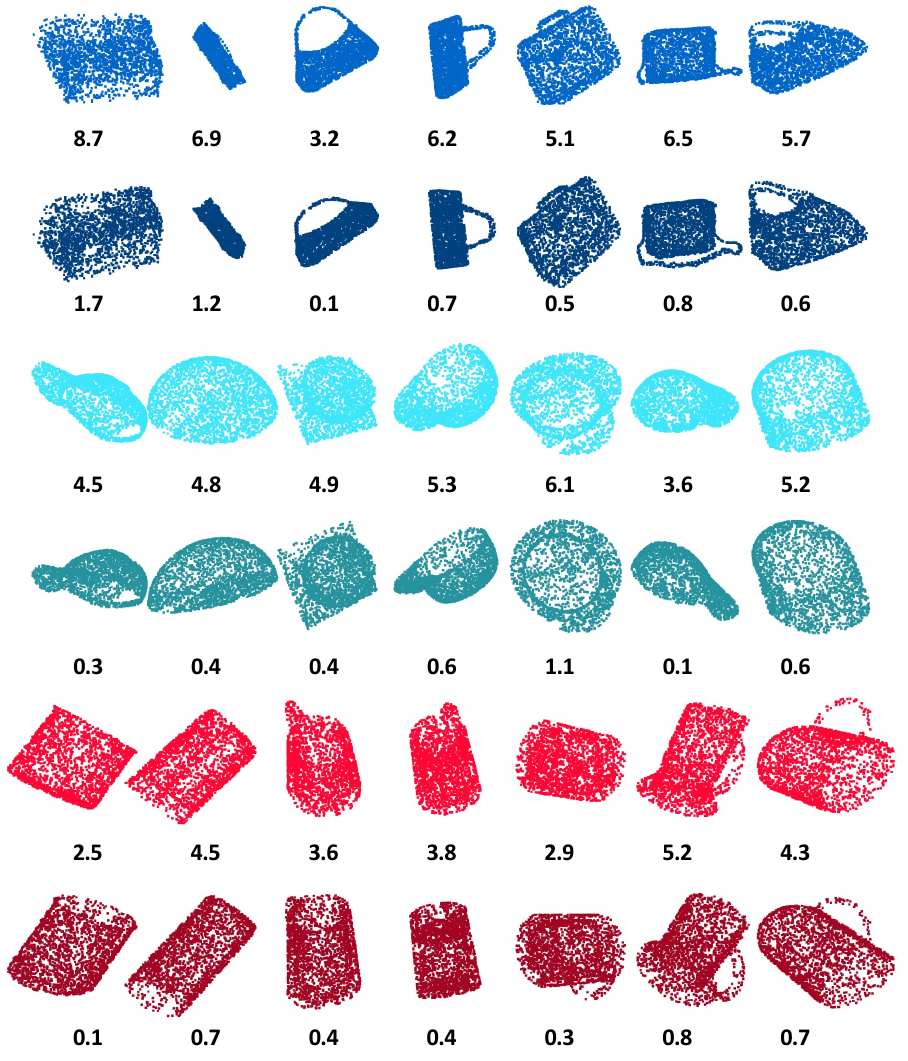}
  \caption{In every two rows, we compare the classification loss of DGCNN \cite{wang2019dynamic} (top row) and ART-DGCNN (bottom row) on the same arbitrarily rotated point clouds, which are randomly sampled from test sets of ShapeNet16 \cite{yi2016scalable}. From top to bottom, the categories of point clouds are ``bag'', ``cap'' and ``mug''.}
  \label{fig4}
\end{figure*}

\section{Visualization}
\label{s6}
Finally, we compare the classification loss of different models under the the randomly rotated test set of ModelNet40 \cite{wu20153d} (Fig. \ref{fig3}) and ShapeNet16 \cite{yi2016scalable} (Fig. \ref{fig4}). 
We illustrate the corresponding loss value under each rotated sample and compare them between the original DGCNN \cite{wang2019dynamic} and our best model ART-DGCNN. 
As can be seen, our method generally shows lower classification loss under both randomly rotated datasets.

\end{document}